\documentclass[letterpaper]{article} 
\usepackage{aaai24}  
\usepackage{times}  
\usepackage{helvet}  
\usepackage{courier}  
\usepackage[hyphens]{url}  
\usepackage{graphicx} 
\usepackage{natbib}  
\usepackage{caption} 
\usepackage{algorithm}
\usepackage{algorithmic}

\usepackage{newfloat}
\usepackage{listings}
\DeclareCaptionStyle{ruled}{labelfont=normalfont,labelsep=colon,strut=off} 
\floatstyle{ruled}
\newfloat{listing}{tb}{lst}{}
\floatname{listing}{Listing}
\usepackage{graphicx}
\usepackage{amsmath}

\DeclareMathOperator*{\argmin}{arg\,min}
\usepackage{multirow}
\usepackage{mathtools}
\DeclarePairedDelimiter{\norm}{\lVert}{\rVert_2}

\usepackage{makecell, booktabs}

\usepackage{xcolor}
\usepackage{CJKutf8}
\newcommand\cn[1]{\begin{CJK*}{UTF8}{gbsn}#1\end{CJK*}}

\title{Machine-Created Universal Language for Cross-lingual Transfer}
\author{
    Yaobo Liang, Quanzhi Zhu, Junhe Zhao, Nan Duan
}
\affiliations{
    Microsoft Research Asia\\
    \{yaobo.liang, v-quanzhizhu, v-junhezhao, nanduan\}@microsoft.com

}

\usepackage{bibentry}

\begin{document}

\maketitle

\begin{abstract}
There are two primary approaches to addressing cross-lingual transfer: multilingual pre-training, which implicitly aligns the hidden representations of various languages, and translate-test, which explicitly translates different languages into an intermediate language, such as English. Translate-test offers better interpretability compared to multilingual pre-training. However, it has lower performance than multilingual pre-training~\cite{conneau2019xlm, xlmr} and struggles with word-level tasks due to translation altering word order. As a result, we propose a new Machine-created Universal Language (MUL) as an alternative intermediate language. MUL comprises a set of discrete symbols forming a universal vocabulary and a natural language to MUL translator for converting multiple natural languages to MUL. MUL unifies shared concepts from various languages into a single universal word, enhancing cross-language transfer. Additionally, MUL retains language-specific words and word order, allowing the model to be easily applied to word-level tasks. Our experiments demonstrate that translating into MUL yields improved performance compared to multilingual pre-training, and our analysis indicates that MUL possesses strong interpretability. The code is at: https://github.com/microsoft/Unicoder/tree/master/MCUL.
\end{abstract}

\section{Introduction}
Cross-lingual transfer aims to tackle NLP tasks in multiple languages using training data from only one or a few languages, such as English. There are two primary approaches to addressing cross-lingual transfer: first, multilingual pre-training involves constructing a multilingual encoder, fine-tuning it in English, and directly testing it in other languages. The multilingual encoder combines words from all target languages to create a large vocabulary, and the hidden representations in the intermediate layers are implicitly aligned to facilitate the cross-lingual transfer. Second, the translate-test approach translates the test set of other languages into an intermediate language, typically English. This allows the model to use English as input for both training and testing, explicitly solving cross-lingual tasks.

Compared to multilingual pre-training, translate-test offers better interpretability by utilizing an intermediate language. However, it has two drawbacks: Translate-test yields worse performance compared to cross-lingual transfer. For instance, its performance on XNLI is 3.1\% lower than that of multilingual pre-training~\cite{xlmr}. Translate-test cannot be applied to word-level tasks such as sequential labeling or machine reading comprehension, as translation alters the word order.

To retain the interpretability of an intermediate language while addressing its limitations, we propose to create a new language specifically designed for cross-lingual tasks. This language, created by machines without requiring human expertise, is called the Machine-created Universal Language (MUL). MUL consists of a set of discrete symbols that form a universal vocabulary and an NL-MUL translator for converting multiple natural languages (NL) to MUL. The NL-MUL translator maps shared concepts from different languages to the same universal words, facilitating better cross-lingual transfer. Additionally, it preserves word order and language-specific vocabulary, allowing for easy application to word-level tasks. This is consistent with the research presented by \citealp{chai2022cross}, which indicates that word order does not affect cross-lingual abilities, thus allowing for the preservation of distinct word orders in different languages. To solve cross-lingual NLP tasks, we can translate both the English training dataset and the multilingual test dataset into MUL, enabling the model to use MUL as input for both training and testing.


To create MUL, our approach consists of three components: First, we pre-train the encoder using multilingual MLM loss and generate word alignment supervision on bilingual data, with the word alignment supervision being created through an unsupervised method. Second, we employ an inter-sentence contrastive learning approach to further enhance the alignment of contextualized word embeddings across languages. Lastly, we introduce vector quantization with cross-lingual alignment (VQ-CA) to improve the interpretability of the universal vocabulary.

We conduct experiments on XNLI, NER, MLQA, and Tatoeba using MUL as input. Compared to the combined vocabulary in multilingual pre-training, our model has a smaller vocabulary size and necessitates fewer parameters at the word embedding layer. We obtain comparable results to XLM-R with 50\% fewer parameters and achieve superior results after redistributing the parameters from word embedding to the transformer's weights. Further analysis reveals that MUL exhibits strong interpretability, as translating to MUL results in less ambiguity compared to translating to English.


Our work offers two significant contributions. First, we introduce a new universal language, MUL, along with a translator between multiple natural languages and MUL. Our experiments demonstrate that translating to MUL achieves strong cross-lingual transfer performance and exhibits good interpretability. Second, we propose an innovative approach to create MUL, which incorporates inter-sentence contrastive learning and vector quantization with cross-lingual alignment.

\section{Related Work}
Multilingual pre-training was first proposed by mBERT~\cite{devlin2019mbert}, which extended the pre-training of BERT~\cite{devlin2019bert} to 100 languages by creating a large vocabulary for all languages and building a multilingual encoder. To improve the cross-lingual transfer performance, lots of works extended monolingual pre-training methods to multiple languages and achieved good cross-lingual performance, such as XLM-Roberta~\cite{xlmr} extended Roberta~\cite{liu2019roberta}, mT5~\cite{xue2021mt5} extended T5~\cite{raffel2020t5}, XLM-E~\cite{chi2022xlme} extended Electra~\cite{clark2020electra}. These methods can be improved by introducing bilingual data~\cite{conneau2019xlm,huang2019unicoder} or multilingual knowledge ~\cite{jiang2022xlm} to improve the implicitly cross-lingual alignment between different languages. All of these works take natural language as input and achieve cross-lingual transfer by implicitly cross-lingual alignment. Translate-test is a baseline of XNLI proposed by ~\citealp{conneau2018xnli}. Further experiments show that both XLM ~\cite{conneau2019xlm} and XLM-R~\cite{xlmr} can achieve better performance compared to the translate-test baseline. Our work achieves better performance compared to XLM-R by translating all data to MUL.

Abstract Meaning Representation(AMR)~\cite{banarescu2013abstract} targets to map natural language sentence to abstract graph, and can server as the transfer layer in MT system~\cite{xue-etal-2014-interlingua}. Our work share the same motivation and propose new methods for cross-lingual pre-training.

VQ-VAE is proposed by ~\citealp{vqvae} to create discrete symbols in the neural network, which is usually used to create discrete symbols for image~\cite{ramesh2021dalle, esser2021vqgan}, video~\cite{wu2022nuwa} and audio ~\cite{baevski2020wav2vec}. It's rare to be applied to natural language which is already discrete symbols. The symbols in our MUL have better interpretability than the symbols for other modalities.

\section{Methodology}
In this section, we begin by defining the Machine-created Universal Language (MUL) and providing an overview of its creation process. Following that, we present the detailed steps involved in creating MUL, including multilingual masked language modeling (MLM), inter-sentence contrastive learning, and vector quantization with cross-lingual alignment.

\subsection{Machine-Created Universal Language (MUL)}
MUL comprises a set of discrete symbols that form a universal vocabulary, along with an NL-MUL translator and a MUL-NL translator for translating between multiple natural languages and MUL.

Each symbol in the universal vocabulary is defined as a universal word. Each universal word corresponds to a concept identified by the model. Most universal words can be aligned with words in multiple natural languages, explicitly facilitating cross-lingual transfer. Some universal words correspond to specific words in certain languages, helping to understand linguistic features unique to those languages.

The NL-MUL translator aims to translate natural languages into MUL. It preserves the word order and generates one universal word for each natural word, which assists the model in solving word-level tasks such as sequential tagging and machine reading comprehension. The mapping relationship between natural words and universal words is context-dependent, meaning a single natural word may correspond to different universal words in varying contexts. Therefore, the translation from NL to MUL involves word disambiguation, which can reduce the model's difficulty in accomplishing specific tasks. The MUL-NL translator, on the other hand, restores NL from MUL and calculates the auto-encoder loss during the MUL creation process.

When addressing cross-lingual NLP tasks, we can employ the NL-MUL translator to convert both the English training dataset and the multilingual test dataset into MUL, which can then be used as input for the model.

\begin{figure*}
    \centering
    \includegraphics[width=\textwidth]{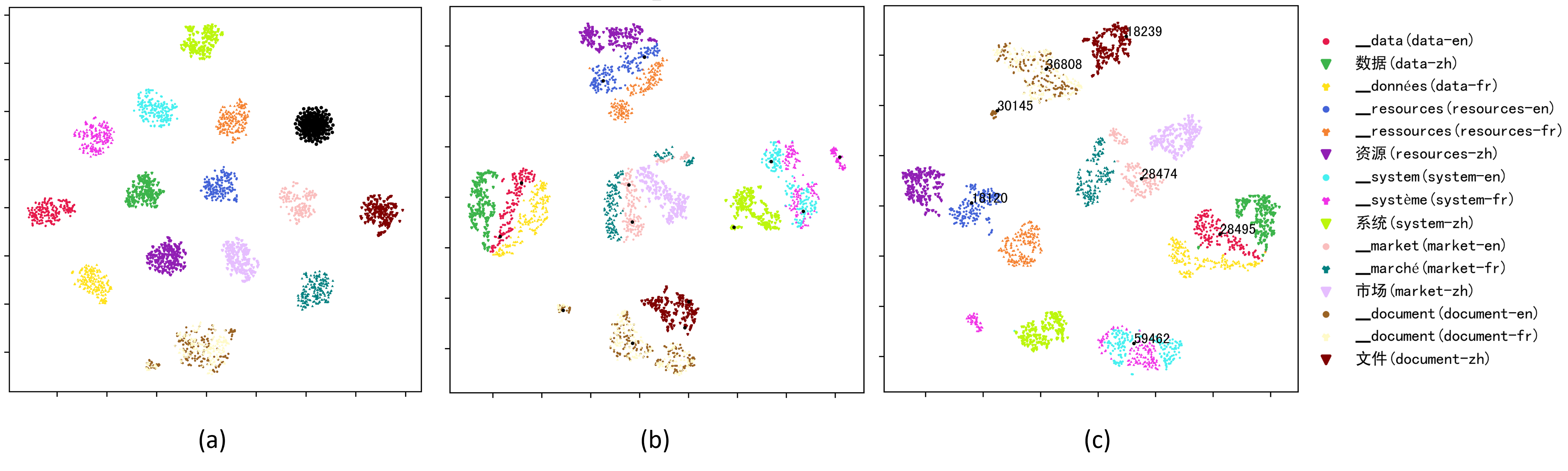}
    \caption{The visualization of contextualized word embeddings at various training stages. Each color represents a word and each point denotes the contextualized embedding of that word in different contexts. Figure 1.a displays the embeddings after pre-training with multilingual MLM, Figure 1.b exhibits the embeddings after inter-sentence contrastive learning, and Figure 1.c demonstrates the embeddings following VQ-CA.} 
    \label{fig:embedding_space}
\end{figure*}
\subsection{Overview of MUL Training}
In order to create MUL, we initially construct an encoder capable of generating contextualized word embeddings for each sentence. For two words with context in different languages, their embeddings are close to one another if and only if they share the same meaning. Subsequently, we create discrete symbols in the embedding space, to ensure that each symbol corresponds to a single concept.

Our approach comprises three components, and we demonstrate their impact on the embedding space with Figure~\ref{fig:embedding_space}. First, we pre-train the encoder using a multilingual masked language model (MLM) loss. The embeddings are depicted in Figure~\ref{fig:embedding_space}.a. Although different words with similar meanings do not have similar embeddings, the encoder can be employed to create unsupervised word alignment labels for bilingual sentence pairs~\cite{awesome}.

Second, we implement an inter-sentence contrastive learning approach to enhance the alignment of contextualized word embeddings across languages. The results can be observed in Figure~\ref{fig:embedding_space}.b, which shows that different words with the same meanings now have similar embeddings.

Lastly, we introduce vector-quantization with cross-lingual alignment (VQ-CA) to establish the universal word list in the universal language. Figure\ref{fig:embedding_space}.b and Figure\ref{fig:embedding_space}.c represent training without and with VQ-CA, respectively. The black points in these figures are the embeddings of the created universal words. For each group of words with the same meanings, we observe that the model trained without VQ-CA generates multiple universal words, while the model trained with VQ-CA produces a single universal word in most instances.

\subsection{Creating Word Alignment Supervision by Multilingual MLM}
First, we pre-train our encoder $Encoder(x)$ using a multilingual Masked Language Model (MLM). This encoder has a vocabulary that includes words from all target languages, as well as a transformer encoder comprising 12 layers.

The contextualized word embeddings generated by the pre-trained encoder demonstrate good performance on the word alignment task~\cite{awesome}. Specifically, the word alignment task involves processing two sentences, $S_s$ and $S_t$, from different languages that have the same meanings. These sentences consist of $n$ and $m$ tokens, respectively, which can be represented as $S_s={s_1, s_2, ..., s_n}$ and $S_t={t_1, t_2, ..., t_m}$. The model's objective is to identify the aligned words or phrases between these two sentences.

We input the two sentences into the pre-trained model $Encoder(x)$ to obtain their contextualized representations, $H_s=Encoder(S_s)={h_{s_1}, h_{s_2}, ..., h_{s_n}}$ and $H_t=Encoder(S_t)={h_{t_1}, h_{t_2}, ..., h_{t_m}}$. The alignment matrix is then computed by $A=H_s H_t^T$.

Next, we apply the softmax function to the first and second dimensions to obtain $A_{t2s}$ and $A_{s2t}$, respectively. The word alignment results are determined by $P=A_{t2s}>c\ \land\ A_{s2t}>c$, where $c$ represents the threshold. Intuitively, this approach identifies the most similar words in the $S_t$ sentence for each word $s_i$ in $S_s$ and vice versa. If both $s_i$ and $t_j$ are the most similar words to each other, they are predicted to be aligned words.

\subsection{Inter-sentence Contrastive Learning}


While the pre-trained contextualized word embeddings can achieve good cross-lingual word alignment performance, there are still two notable shortcomings. Firstly, the distance between aligned words is not close to zero, even though they are the most similar words between the source and target sentences, as illustrated in Figure\ref{fig:embedding_space}.a. Secondly, the distance between words of the same type is too close, and it becomes even closer when the model is trained with vanilla contrastive loss. For instance, words of the same type can include time-related terms such as "year", "month", "day", and "hour", or adverbs of frequency like "always", "never", and "sometimes". In bilingual sentences, there is typically only one or a few words for each type. Consequently, being adept at identifying words of the same type is sufficient for achieving good word alignment performance. However, such granularity is too coarse for MUL. Additional examples and analysis can be found in Appendix A.



To address this issue, we propose inter-sentence contrastive learning. This approach has two main steps. First, we employ contrastive learning to minimize the distance between aligned words while maintaining a larger distance between non-aligned words. Second, we utilize words from other sentence pairs as negative samples to ensure that words of the same type remain distant from one another.



In the contrastive learning process, we consider all aligned words in matrix $P$ as positive pairs. We perform post-processing on the unaligned words in $P$ and represent the negative matrix as $N\in {0,1}^{n\times m}$. Further details can be found in Appendix A. The contrastive loss is defined as 

\begin{align*}
    lo&ss_{cts} = - log \sum_{i,j} P_{ij}\exp{H_{s_i}H_{t_j}^T} \\
    &+ log\sum_{i,j} \left({P_{ij}\exp{H_{s_i}H_{t_j}^T} + N_{ij}\exp{H_{s_i}H_{t_j}^T}}\right)
\end{align*}

In inter-sentence contrastive learning, we sample multiple bilingual pairs and generate a new pair by concatenating the source and target sentences, respectively. For example, consider two pairs: $(S_s^1, S_t^1)$ and $(S_s^2, S_t^2)$. The new pair is $([S_s^1, S_s^2], [S_t^1, S_t^2])$.  We create the positive alignment matrices $P^1$ and $P^2$ for the two pairs separately. Subsequently, we merge the two positive alignment matrices and construct a positive matrix for the concatenated sentence pair: 

\[
P_{inter}=
  \begin{bmatrix}
    P^1 & 0 \\
    0 & P^2 
  \end{bmatrix}
\]

This means that we won't treat any pairs between $(S_s^1, S_t^2)$ and $(S_s^2, S_t^1)$ as positive alignment. We avoid concatenating the two sentences initially to generate $P_{inter}$ directly, as this could introduce additional interference in word alignment and diminish alignment quality. By employing this method, we can effectively push words of the same type further apart. For negative pairs, we apply the same post-processing technique.

\subsection{Vector Quantization with Cross-lingual Alignment (VQ-CA)}
To create a universal vocabulary, one option is using VQ-VAE \cite{vqvae} to learn a set of discrete symbols. However, the symbols generated by VQ-VAE lack clear meanings and are difficult for humans to comprehend. For instance, in Figure\ref{fig:embedding_space}.b, multiple symbols are created for each meaning, and each symbol lacks a precise definition. So we propose Vector-Quantization with Cross-Lingual Alignment (VQ-CA) to guide the learning of discrete symbols by aligning them with multiple languages simultaneously. In most cases, the symbols produced by VQ-CA correspond to a single concept, making them easier to understand compared to those created by VQ-VAE.

We define the embedding of universal vocabulary as $e=\{e_1, e_2, ..., e_K\}$, where $e_i\in R^D$ is the embedding of discrete symbol $i$. $K$ is the size of the universal vocabulary and $D$ is the dimension of hidden representation. 

Our model comprises an $Encoder(x)$ and a $Decoder(x)$. The $Encoder(x)$ contains word embedding layers and multiple transformer layers. For a sentence $S={s_1, s_2, ..., s_n}$, we map it to contextualized word embeddings $H=Encoder(S)={h_{1}, h_{2}, ..., h_{n}}$. We generate the sentence in the universal language by mapping each contextualized word representation $h_{i}$ to symbol $k_{i}=Quantize(h_{i})=\argmin_j\norm{e_j-h_{i}}$. The sentence in MUL is $S_u={k_{1}, k_{2}, ..., k_{n}}$ and its embedding is $E={e_{k_{1}}, e_{k_{2}}, ..., e_{k_{n}}}$. The $Encoder(x)$ and $Quantize(x)$ together form the NL-MUL translator. The $Decoder(x)$ consists of several transformer layers and a softmax layer, which can generate the probability of mapping the sentence embedding in MUL back to natural language as $P(S|E)=Decoder(E)$. The $Decoder(x)$ serves as the MUL-NL translator.

To train the $Encoder(x)$, $Decoder(x)$ and universal vocabulary embedding $e$, our loss is:

\begin{align*}
    loss_{VQ-CA} =&\; log P(S|E) + \norm{sg(E)-H} \\
    &+ \lambda \norm{E-sg(H)} + loss_{CA}
\end{align*}

The notation $sg(x)$ represents the stop gradient operation. The first three losses are derived from VQ-VAE. The first term is the auto-encoder loss, which aims to recover the original natural language sentences from the MUL sentences. The second term constraint contextualized word embeddings to be close to universal language embeddings. In our experiment  The third term constraints universal language embeddings to be close to contextualized word embeddings. In our experiments, we find that the update speed for the embeddings of the universal language is too slow. Consequently, we replace the third loss with exponential moving averages, following the approach of ~\citealp{vqvae}.


The fourth loss, $L_{CA}$, constrains the aligned words to map to the same symbol. For aligned words that map to different symbols, the loss $L_{CA}$ pushes one symbol away and retains only the other symbol in the nearby region. Consequently, both words can be mapped to the preserved symbol, ensuring that aligned words share the same symbol in the universal language representation.

\begin{figure}[ht]
    \centering
    \includegraphics[width=0.45\textwidth]{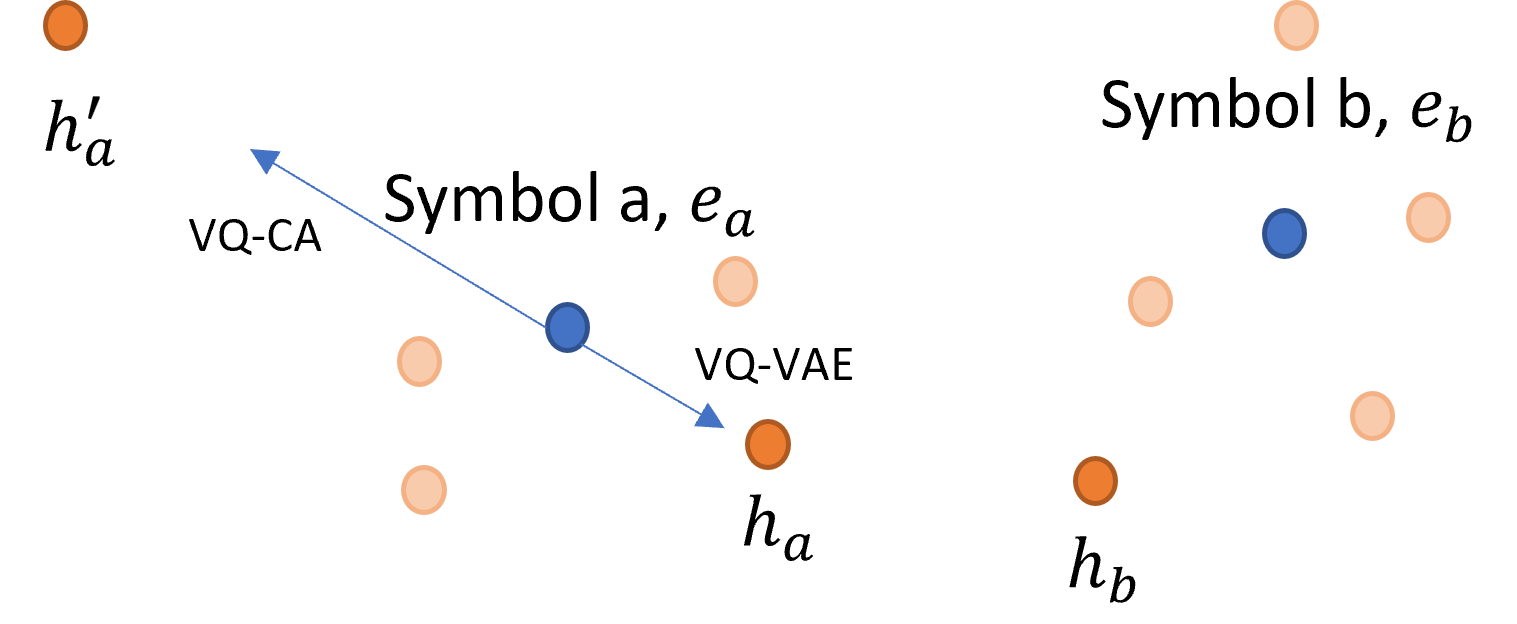}
    \caption{Visualization of VQ-CA. The orange dots shows the embeddings related to a pair of aligned words. The light orange dots shows the embeddings that map to symbol a and b.}
    \label{fig:push}
\end{figure}

We illustrate $loss_{CA}$ in Figure~\ref{fig:push}. Formally, let's consider two aligned words with embeddings $h_a$ and $h_b$. We quantize them to two symbols $a=quantize(h_a)$ and $b=quantize(h_b)$. The original VQ-VAE loss requires the symbol $a$ to move towards $h_a$ and symbol $b$ to move towards $h_b$. However, in $loss_{CA}$, one symbol should be pushed away. The selection of which symbol to push away is determined by the number of natural language words that are mapped to it. Without loss of generality, we assume that $a$ should be pushed away. Then we create an embedding $h_{a}'=e_a+\lambda * (e_a-h_a)$ at the opposition direction of $h_a$, and add a new loss $loss_{CA}=\norm{e_a-h_{a}'}$ for it. We also use exponential moving averages to update $e_a$. This loss moves $e_a$ in the direction opposite to that of the VQ-VAE. Once $a$ has been moved far away from $h_a$, the nearest symbol of $h_a$ may change to $b$. As training progresses, symbol $b$ will dominate the region of symbols $a$ and $b$, while symbol $a$ will fade away.

\begin{table*}[t]
        \centering
        \resizebox{1\linewidth}{!}{
    \begin{tabular}{cc|cccccccccccccccc}
    \hline
Model & Parameter & en & de & fr & es & el & bg & ru & tr & ar & vi & th & zh & hi & sw & ur & avg \\
 \hline
mBERT & 178M & 82.1 & 73.8 & 74.3 & 71.1 & 66.4 & 68.9 & 69 & 61.6 & 64.9 & 69.5 & 55.8 & 69.3 & 60.0 & 50.4 & 58.0 & 66.3 \\
XLM & 250M & 85.0 & 78.7 & 78.9 & 77.8 & 76.6 & 77.4 & 75.3 & 72.5 & 73.1 & 76.1 & 73.2 & \textbf{76.5} & 69.6 & 68.4 & 67.3 & 75.1 \\
XLM-R Base & 278M & 85.3 & 78.3 & 79.2 & 79.9 & 77.3 & 78.6 & 76.1 & 74.7 & 73.8 & 75.6 & \textbf{73.3} & 74.6 & 71.7 & 68.6 & 68.2 & 75.7 \\
mT5 Base & 580M & 84.7 & 77.4 & 79.1 & 80.3 & 77.1 & 78.6 & 77.1 & 72.8 & 73.3 & 74.2 & 73.2 & 74.1 & 70.8 & 69.4 & 68.3 & 75.4 \\
Unicoder & 278M & 85.4 & 78.2 & 79.2 & 79.8 & 77.3 & 78.5 & 76.7 & 73.8 & 73.9 & 75.9 & 71.8 & 74.7 & 70.1 & 67.4 & 66.3 & 75.3 \\
InfoXLM & 278M & 86.4 & 79.3 & 80.3 & 80.9 & 77.8 & 79.3 & 77.6 & 75.6 & 74.2 & 77.1 & 74.6 & 77.0 & 72.2 & 67.5 & 67.3 & 76.5 \\
MUL Small & 132M & 84.0 & 78.5 & 79.5 & 79.9 & 78.4 & 79.0 & 75.8 & 74.4 & 74.8 & 75.8 & 70.9 & 73.8 & 70.8 & 71.1 & 68.1 & 75.7 \\
MUL Base & 277M & \textbf{85.5} & \textbf{80.5} & \textbf{81.1} & \textbf{81.4} & \textbf{79.8} & \textbf{80.6} & \textbf{78.4} & \textbf{75.9} & \textbf{77.4} & \textbf{78.4} & 72.8 & 76.0 & \textbf{73.8} & \textbf{72.9} & \textbf{69.9} & \textbf{77.6} \\
\hline
    \end{tabular}
    }
    \caption{Evaluation results on XNLI.}
    \label{tab:xnli}
\end{table*}

\begin{table}[h]
        \centering
        \small
    \begin{tabular}{c|cccccc}
    \hline
Model &  NER & MLQA & Tatoeba \\
 \hline
XLM-R Base & 61.9 & 65.6 / 47.9 & 63.4 \\
mT5 Base & 59.5 & 64.4 / 45.0 & - \\
InfoXLM & - & 68.1 / 49.7 & 77.8 & \\
MUL Small & 60.8 & 65.6 / 47.4  & 74.6 \\
MUL Base & \textbf{63.0} & \textbf{69.4 / 50.8} & \textbf{79.3} \\
\hline
    \end{tabular}
    \caption{Evaluation results on three cross-lingual tasks.}
    \label{tab:five_tasks}
\end{table}

\section{Experiments}
In this section, we begin by presenting the training details, followed by experiments on four diverse cross-lingual tasks. Lastly, we conduct the ablation study to examine the different components of our method.

\begin{table*}
        \centering
        \small
    \begin{tabular}{c|ccc|c}
    \hline
Setting & Precision $\uparrow$ & Recall $\uparrow$ & AER $\downarrow$ & XNLI $\uparrow$ \\
\hline
MUL (pair=4) & 90.0 & \textbf{51.2} & \textbf{35.3} &  \textbf{74.0} \\
w/o VQ-CA & 89.1 & 42.3 & 43.4 &  73.7 \\
w/o contrastive loss + VQ-CA  & 69.2 & 11.5 & 80.9 & 69.7 \\
w/o inter-sentence contrastive loss (pair=1) & 90.1 & 46.0 & 39.8 & 72.9 \\
inter-sentence contrastive loss (pair=2) & \textbf{90.5} & 49.5 & 36.6 & 73.4 \\
\hline
    \end{tabular}
    \caption{The ablation study of MUL. The first row is the best setting in our paper which uses inter-sentence contrastive on 4 pairs of sentences. We skip pre-training and only fine-tuning in these experiments to reduce computational costs.}
    \label{tab:ablation}
\end{table*}

\begin{table*}[t]
    \centering
    \small
    \begin{tabular}{c|p{13cm}}
    \hline
    Language & \multicolumn{1}{c}{Sentence in Natural Language and MUL} \\
    \hline 
    \multirow{2}{*}{English} & He is sitting in a chair \\
    & \_He/\colorbox{red!60}{45816} \_is/\colorbox{yellow!60}{36575} \_sitting/\colorbox{green!60}{44530} \_in/43023 \_a/\colorbox{violet!30}{29017} \_chair/\colorbox{blue!40}{43227} \\[0.2cm]
    \multirow{2}{*}{French} & Il est assis sur une chaise \\
    & \_Il/\colorbox{red!60}{45816} \_est/55230 \_assis/\colorbox{green!60}{44530} \_sur/7419 \_une/\colorbox{violet!30}{29017} \_chaise/\colorbox{blue!40}{43227} \\[0.2cm]
    \multirow{2}{*}{Chinese} & \cn{他正坐在椅子上} \\
    & \cn{\_他/\colorbox{red!60}{45816} 正/\colorbox{yellow!60}{36575} 坐在/\colorbox{green!60}{44530} 椅/\colorbox{blue!40}{43227} 子/32533 上/17777} \\
    \hline \multirow{2}{*}{English} & She serves as the chair in our committee \\
    & \_She/\colorbox{magenta!60}{28168} \_serves/13352 \_as/45554 \_the/\colorbox{cyan!80}{50140} \_chair/\colorbox{lime}{38789} \_in/43267 \_our/\colorbox{teal!60}{10816} \_committee/\colorbox{brown!60}{59378} \\[0.2cm]
    \multirow{2}{*}{French} & Elle est la présidente de notre comité \\
    & \_Elle/\colorbox{magenta!60}{28168} \_est/22312 \_la/\colorbox{cyan!80}{50140} \_présidente/\colorbox{lime}{38789} \_de/29699 \_notre/\colorbox{teal!60}{10816} \_comité/\colorbox{brown!60}{59378} \\[0.2cm]
    \multirow{2}{*}{Chinese} & \cn{她在我们的委员会中担任主席} \\
    & \cn{\_她/\colorbox{magenta!60}{28168} 在/50000 我们的/\colorbox{teal!60}{10816} 委员会/\colorbox{brown!60}{59378} 中/10923 担任/43518 主席/\colorbox{lime}{38789}} \\
    \hline
    \end{tabular}
    \caption{The examples to translate natural language sentences into universal language. For each example, we show the results of tokenization and the universal word corresponding to each token.}
    \label{tab:examples_chair}
\end{table*}

\begin{table*}
    \centering
    \small
    \begin{tabular}{c|ccc}
         \hline
         \thead{Universal Word} & English Word & French Word & Chinese Word \\
         \hline
\colorbox{blue!40}{43227} & \_chair:46\%, \_wheelchair:33\% & \_fauteuil:44\%, \_chaise:35\% & \cn{椅:95\%, 凳:3\%} \\
\colorbox{lime}{38789} & \_Chair:40\%, \_Chairman:28\% & \_président:48\%, président:21\% & \cn{主席:98\%, 院长:1\%} \\
\colorbox{yellow!60}{36575} & \_is:68\%, s:32\% & \_est:49\%, \_a:18\% & \cn{正在:64\%, 正:22\%} \\
\colorbox{violet!30}{29017} & \_a:92\%, \_an:7\% & \_un:37\%, \_une:28\% & \cn{一个:69\%, 一:11\%} \\
\colorbox{red!60}{45816} & \_He:94\%, \_he:4\% & \_Il:89\%, \_il:5\% & \cn{\_他:95\%, \_He:2\%} \\
\colorbox{magenta!60}{28168} & \_She:77\%, \_she:21\% & \_Elle:76\%, \_elle:19\% & \cn{\_她:81\%, 她:17\%} \\
\colorbox{teal!60}{10816} & \_our:100\%, \_Our:0\% & \_notre:57\%, \_nos:41\% & \cn{我们:81\%, 我国:9\%} \\
         \hline
    \end{tabular}
    \caption{For each universal word, we list the top 2 natural words that correspond to it in three languages.}
    \label{tab:code_means_chair}
\end{table*}

\subsection{Training Details}
In the first stage, we pre-train the encoder with a multilingual MLM objective on 15 languages of XNLI. The vocabulary size is 250K, and the model contains 12 layers and 768 hidden states, identical to XLM-R base. Limited by resources, we pre-train the model for 500K steps with a batch size of 8192, which is less than XLM-R Base. The pre-training corpus is CC-Net ~\cite{ccnet}.

In the second stage, we train our model on bilingual data OPUS-100 ~\cite{zhang2020improving}. The encoder has 8 layers, and the decoder has 4 layers. They are initialized by the first 8 and last 4 layers of the encoder pre-trained in the first stage. We select 8 as encoder layers because previous research\cite{awesome} shows that outputs of the 8th layer have the best cross-lingual alignment quality. The size of the universal vocabulary $K$ is set to 60K, as the vocabulary size of GPT is 50K.

Once we have the encoder, decoder, and universal vocabulary, we proceed with pre-training and fine-tuning on MUL. As both pre-training and fine-tuning require multiple epochs, we translate the corpus into MUL during the pre-processing stage, saving significant time. The vocabulary size is reduced from 250K to 60K. We try two sets of model sizes: MUL Small and MUL Base. The small model has the same layer number and hidden size as XLM-R Base, with the total parameter number being only half of XLM-R Base, due to the reduction in vocabulary size. The base model reallocates parameters from embedding layers to transformer layers, keeping the total parameter number unchanged. The details are listed in Appendix B. The hyper-parameters in pre-training and fine-tuning are the same as those of natural language. We run all fine-tuning experiments four times and report the average of the results.

\subsection{Performance on Cross-lingual Tasks}
We test MUL on four diverse cross-lingual tasks: cross-lingual Natural Language Inference (XNLI)\cite{conneau2018xnli} is a sentence classification task; NER\cite{Pan2017ner} is a sequential labeling task; MLQA \cite{Lewis2020mlqa} is a machine reading comprehension task; Tatoeba \cite{Artetxe2019tatoeba} is a cross-lingual sentence retrieval task. We only use English training data in the first three tasks and don't use any training data in Tatoeba.


We compare our model with six baseline models that use natural language as input. The first three models are pre-trained exclusively on monolingual datasets: mBERT \cite{devlin2019mbert} and XLM-R Base \cite{xlmr} share the same pre-training objective as ours, while mT5 \cite{xue2021mt5} is pre-trained using a denoising objective. The last three models are pre-trained on both monolingual and bilingual datasets: XLM \cite{conneau2019xlm} employs multilingual MLM and TLM in the 15 languages of XNLI. Unicoder \cite{liang2020xglue} and InfoXLM \cite{chi2021infoxlm} introduce new bilingual objectives; their monolingual datasets are the same as our model, but their bilingual datasets are larger. For a fair comparison, we continue to pre-train XLM-R Base on the 15 languages of XNLI.

We show the performance of XNLI for each language in Table \ref{tab:xnli}, and present the results on NER, MLQA, and Tatoeba in Table \ref{tab:five_tasks}.
Based on the results, we can draw three conclusions: 1) MUL Base achieves the best performance on all tasks with the same parameter number as XLM-R Base, Unicoder, and InfoXLM. This demonstrates that taking MUL as input can achieve excellent cross-lingual transfer performance. 2) MUL Small also achieves comparable performance to baselines with minimal parameters. On Tatoeba, it achieves better performance compared to XLM-R Base and slightly lower than InfoXLM, which introduces sentence-level contrastive learning. On XNLI, MLQA, and NER, MUL Small can achieve comparable results to baselines. 3) On XNLI, both MUL Small and Base achieve good performance on low-resource languages, such as Swahili (sw) and Urdu (ur).


\subsection{Ablation Study}
We evaluate the quality of MUL using performance on word alignment and XNLI:

\textbf{Word alignment with MUL} We translate natural language sentences into MUL and predict aligned words by checking if they correspond to the same universal word. Word alignment can help us understand whether words with the same meanings are mapped to the same universal word. We report three metrics: precision, recall, and alignment error rate (AER). We don't train our model on the word alignment training dataset and directly evaluate it on the test dataset. We evaluate our model on German-English (de-en), French-English (fr-en), and Chinese-English (zh-en) and report the averaged results. The test datasets come from ~\citealp{mihalcea2003evaluation, vilar2006aer, liu2015contrastive} respectively.


\textbf{XNLI results} We report the results on XNLI to evaluate the quality of using MUL as input to solve cross-lingual tasks. We don't conduct pre-training on MUL in the ablation study limited by resources. In fine-tuning, we load the transformer weight of the pre-trained encoder. The word embedding of each universal word is the weighted sum of its corresponding natural words, and the weights are the frequency of the corresponding natural words.


The results of the ablation study are presented in Table \ref{tab:ablation} and include two aspects: 

\textbf{Ablation of loss} After removing VQ-CA, the recall of word alignment drops about 10 percent. This is because the model without VQ-CA often generates multiple universal words for the same concept. As a result, aligned words are mapped to different concepts even if they have similar contextualized word embeddings. After removing both of them, the performance on word alignment becomes very poor, and the performance on XNLI drops significantly. This is because the embeddings of aligned words are far from each other and are mapped to different universal words.

\textbf{Ablation of inter-sentence contrastive learning} The inter-sentence contrastive learning leverages multiple sentence pairs, and we report the performance of 1, 2, and 4 sentence pairs. Using one sentence pair means vanilla contrastive and removes the inter-sentence strategy. We find that a larger number of sentence pairs leads to better performance both on word alignment and XNLI. However, increasing the sentence pair numbers also increases GPU memory usage and training time, so we can only set it to 4.




\section{Analysis}
We conduct the analysis focusing on three aspects: the interpretability of MUL, the word disambiguation in NL-MUL translation, and the language-specific words in MUL.
\begin{table*}[t]
    \centering
    \small
    \begin{tabular}{c|l}
    \hline
    Language & \multicolumn{1}{c}{Sentence in Natural Language and MUL} \\
    \hline 
    \multirow{2}{*}{English} & Do cats eat fish \\
    & \_Do/7221 \_cats/44206 \_eat/49321 \_fish/53877 \\[0.1cm]
    \multirow{2}{*}{Chinese} & \cn{猫吃鱼吗} \\
    & \cn{猫/44206 吃/49321 鱼/53877 吗/28009} \\
    \hline
    \multirow{2}{*}{English} & I am going to play basketball \\
    & \_I/12146 \_am/13086 \_going/48222 \_to/34958 \_play/33633 \_basketball/31719 \\[0.1cm]
    \multirow{2}{*}{Chinese} & \cn{我要去打篮球} \\
    & \cn{\_我/12146 要去/10072 打/27676 篮球/31719} \\
    \hline
    \end{tabular}
    \caption{Examples of language-specific words in MUL that correspond to words in a few natural languages.}
    \label{tab:example_cat_fish}
\end{table*}

\subsection{The Interpretability of MUL}
To better understand MUL, we show two groups of examples in Table \ref{tab:examples_chair}. Each group contains three sentences in English, French, and Chinese, all with the same meanings. We first tokenize these sentences and then translate them into MUL.

To understand the meaning of each universal word, we can summarize the natural words that are often translated into it. In Table \ref{tab:code_means_chair}, we list the top 2 natural words that correspond to the universal word in three languages. For example, the universal word ``43227" corresponds to ``chaise" in French and ``\cn{椅}" in Chinese, which helps us to know that ``43227" means a chair, which is a piece of furniture for one person to sit on. Similarly, we can deduce that ``38789" means the person in charge of the meeting based on ``président" in French.

For most words in different languages with the same meanings, their universal words are the same as each other. By mapping to the same universal words, knowledge can be easily transferred between languages, enabling effective cross-lingual learning and understanding.

\subsection{Language Specific Words in MUL}

MUL maps words with the same meanings from different languages to a single universal word while retaining language-specific words for each language. We present two cases in Table \ref{tab:example_cat_fish}. To transform a declarative sentence into a question, we can add ``do" before the sentence in English. However, in Chinese, a special word ``\cn{吗}" must be added to the end of the sentence. These words do not correspond to any word in their translations, but they play a crucial syntactic role and aid in understanding the meaning of sentences. Additionally, the English phrase ``am going to" consists of three words, while it is aligned to the single word ``\cn{要去}" in Chinese, resulting in their universal words not being identical. We provide more analysis in Appendix C.

MUL retains language-specific words for a better understanding of each language, while translation to any natural language will lose them. Consequently, MUL represents the union of all natural languages and contains richer information than any individual natural language.

\subsection{Word Disambiguation of NL-MUL Translation}
For a word that has different meanings in different contexts, it may correspond to different universal words during the NL-MUL translation. For example, in Table~\ref{tab:examples_chair}, ``chair" means ``furniture" in the first group and means ``a person" in the second group, so it corresponds to different universal words. Compared to natural words, the meaning of universal words is closer to concepts shared across multiple languages. This makes them less ambiguous. For example, we can distinguish the meaning of 43227 and 38789, while we can't distinguish the meanings of two instances of ``chair" without context.

We conduct more statistical experiments on the CoarseWSD-20 dataset~\cite{loureiro2021coarsewsd} and present the results in Table \ref{tab:word_chair}, Table \ref{tab:word_apple}, and Table \ref{tab:word_club}. We can find that the universal words of ``chair" and ``apple" have a good correlation to concepts, while the universal words for ``club" are the same in most cases. This is because most of the ``club" instances in bilingual data correspond to the first concept, only 3\% of ``club" means ``nightclub," and almost no ``club" means ``club (weapon)." This shows that translating to MUL can disambiguate parts of words, but the disambiguation is not good enough due to the unbalanced distribution of concepts in our data.

During translation, two different words in non-English languages may be translated into the same word in English. This increases the ambiguity of words. However, when translated into different universal words, this ambiguity is reduced. By using universal words, the difficulty of solving NLP tasks is decreased.

\begin{table}[h]
    \centering
    \small
    \begin{tabular}{c|cc}
    \hline
       chair & chairman & Chair (seat)\\\hline
30320 & 2 & 0\\
38789 & 100 & 13\\
43227 & 11 & 102\\
53430 & 2 & 0\\\hline
    \end{tabular}
    \caption{The statistics of relation between universal words and different meanings of ``chair".}
    \label{tab:word_chair}
\end{table}
\begin{table}[h]
    \centering
    \small
    \begin{tabular}{c|cc}
    \hline
       apple & apple\_inc & apple (fruit)\\\hline
18766 & 668 & 44\\
20027 & 224 & 848\\\hline
    \end{tabular}
    \caption{The statistics of the relation between universal words and different meanings of ``apple".}
    \label{tab:word_apple}
\end{table}
\begin{table}[h]
    \centering
    \small
    \begin{tabular}{c|ccc}
    \hline
       club & club & nightclub & Club (weapon)\\ \hline
50064 & 54 & 54 & 54\\\hline
    \end{tabular}
    \caption{The statistics of the relation between universal words and different meanings of ``club".}
    \label{tab:word_club}
\end{table}

\section{Conclusion}
In this work, we present a new universal language MUL created by machines, which can serve as an intermediate language and solve cross-lingual tasks by translating all languages into MUL. We introduced inter-sentence contrastive learning and VQ-CA which are critical to creating MUL. The experiments show that the model with MUL as input achieves excellent cross-lingual performance and greatly reduces the size of vocabulary size. Further analysis shows the good interpretability of MUL and the capability for word disambiguation.

\bibliography{aaai24}

\appendix

\section{A: Inter-Sentence Contrastive Learning}
\label{sec:contrastive}

\subsection{Examples of Word-Alignment}
\begin{figure*}[t]
    \centering
    \includegraphics[width=0.9\textwidth]{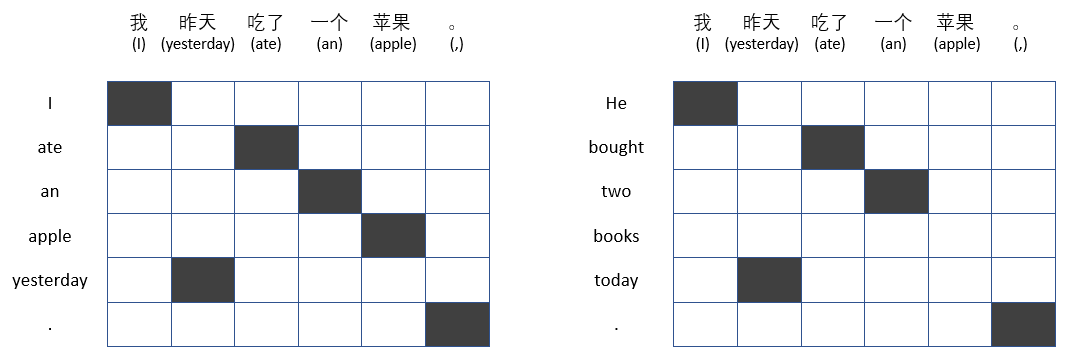}
    \caption{The word alignment results generated by pre-trained contextualized word embeddings. It could align the words with similar but not identical meanings in two sentences. For example, the word "\cn{一个}" in Chinese means "an" or "one" in English. But when the English sentence has another number word "two", the model also will predict that they are aligned with each other.}
    \label{fig:word_alignment}
\end{figure*}

Figure ~\ref{fig:word_alignment}.a shows that the model successfully aligns the correct words. But if we replace the English sentence with a similar sentence and keep the Chinese sentence unchanged, the model still generates lots of alignment pairs which is illustrated in Figure~\ref{fig:word_alignment}.b. Such as both "\cn{我}(I)" and "He" are personal pronouns, and both "\cn{一个}(an)" and "two" are number words. This shows that the model will predict the words with the same type as the positive alignment pair, while it's not necessary to have the same meaning. To generate a better universal language, we hope the words with the same type but not the same meanings don't be close to each other.
\subsection{Post-processing of Negative Pairs in Contrastive Learning}
 For negative pairs $(s_i, t_j)$, we remove the pairs if there is another positive pair $(s_k, t_l)$ where $s_i=s_k$ and $t_j=t_k$. For example for two sentences "He bought two books and two pens" and "Il a acheté deux livres et deux stylos", there are two "two" in $S_t$ and two "deux" in $S_s$. The alignment model will generate two pairs as positive alignment. But we won't treat the pair of first "two" and second "deux" as negative pairs. Because we still hope these two "two" could be mapped to the same universal word although they have a different semantic role in this sentence.

\section{B: The Detailed Hyper-parameters}
In second stage, we train the model for 50K steps with batch 4196. The learning rate is warmed up over the first 1K steps to a peak value of 3e-5, and then linearly decayed. The universal vocabulary embedding is initialized by Gaussian distribution whose mean is 0 and variance is 1.

\label{sec:model_size}
The model sizes are listed in Table \ref{tab:model_size}. The small model has the same layer number and hidden size compared to XLM-R Base. The hidden size of the medium model is 1024, which follows the setting of Bert Large and XLM-R Large. The layer number is 17 so the total parameter number is the same as XLM-R Base. Most of the parameters of XLM-R Base are used by word embeddings to cover large vocabulary from different languages. While our universal language could reduce the vocabulary size dramatically and reallocate the parameter from reduced embedding layers to transformer layers.
\begin{table*}[h]
    \centering
    \begin{tabular}{cccccc}
    \hline
    Model Settings & Vocabulary Size & Hidden Size & \thead{Layer\\ Number} & \thead{Embedding Parameter\\ Number} & \thead{total parameter\\ Number}  \\
    \hline
    XLM-R Base & 250K & 768 & 12 & 192M & 278M \\
MUL Small & 60K & 768 & 12 & 46M & 132M \\
MUL Base & 60K & 1024 & 17 & 61M & 277M \\
\hline
    \end{tabular}
    \caption{The detailed parameters of XLM-R Base and our models.}
    \label{tab:model_size}
\end{table*}



\section{C: Language Specific Words in MUL}
\label{sec:statistics_ul_nl}

For each language, we sample a subset of the corpus and translate it into MUL. We find that tokens from all languages correspond to 42K universal words, while each language corresponds to 27K universal words on average. This indicates that although MUL reduces the vocabulary from 250K to 42K by merging natural words, there are still some unaligned language-specific words. Detailed statistics can be found in table\ref{tab:statistics_each_language}. We report four statistics for universal language: "Universal words/Natural words" represents the number of universal words per natural word mapped to; "Natural words/Universal words" represents the number of natural words per universal word mapped to; "Universal words number" indicates the number of universal words used by the model; and "Natural Words Number" represents the number of natural words in the original corpus.

\begin{table*}[h]
    \centering
    \begin{tabular}{c|cccc}
    \hline
Language & \thead{Universal Words/\\Natural Words} & \thead{Natural Words\\Number} & \thead{Natural words/\\Universal Words} & \thead{Universal Words\\Number} \\
\hline
ar & 1.85 & 76,376 & 4.24 & 27,830 \\
bg & 1.74 & 78,718 & 4.05 & 26,042 \\
de & 1.75 & 70,427 & 3.72 & 26,239 \\
el & 1.76 & 79,016 & 4.10 & 27,156 \\
en & 1.67 & 65,001 & 3.45 & 25,020 \\
es & 1.88 & 70,822 & 3.99 & 26,671 \\
fr & 1.79 & 73,513 & 3.79 & 27,061 \\
hi & 1.69 & 59,766 & 3.17 & 27,319 \\
ru & 1.78 & 75,165 & 4.32 & 25,858 \\
sw & 1.6 & 63,789 & 3.14 & 25,688 \\
th & 1.73 & 76,716 & 3.70 & 30,135 \\
tr & 1.7 & 71,131 & 3.72 & 27,052 \\
ur & 1.73 & 68,418 & 3.23 & 27,194 \\
vi & 1.53 & 60,264 & 2.70 & 27,124 \\
zh & 1.92 & 58,445 & 4.02 & 28,647 \\
all & 2.81 & 249,973 & 14.04 & 42,269 \\
avg & 1.74 & 69,837 & 3.69 & 27,002 \\
\hline
    \end{tabular}
    \caption{The statistics of MUL on corpus from different languages.}
    \label{tab:statistics_each_language}
\end{table*}
\end{document}